# Automatic tuning of communication protocols for vehicular ad hoc networks using metaheuristics

J. García-Nieto, J. Toutouh, E. Alba *

Dept. de Lenguajes y Ciencias de la Computación, University of Málaga, ETSI Informática, Campus de Teatinos, Málaga 29071, Spain

A B S T R A C T

The emerging field of vehicular ad hoc networks (VANETs) deals with a set of communicating vehicles which are able to spontaneously interconnect without any pre-existing infrastructure. In such kind of networks, it is crucial to make an optimal configuration of the communication protocols previously to the final network deployment. This way, a human designer can obtain an optimal QoS of the network beforehand. The problem we consider in this work lies in configuring the File Transfer protocol Configuration (FTC) with the aim of optimizing the transmission time, the number of lost packets, and the amount of data transferred in realistic VANET scenarios. We face the FTC with five representative state-of-the-art optimization techniques and compare their performance. These algorithms are: Particle Swarm Optimization (PSO), Differential Evolution (DE), Genetic Algorithm (GA), Evolutionary Strategy (ES), and Simulated Annealing (SA). For our tests, two typical environment instances of VANETs for Urban and Highway scenarios have been defined. The experiments using *ns-* 2 (a well-known realistic VANET simulator) reveal that PSO outperforms all the compared algorithms for both studied VANET instances.

## 1. Introduction

Vehicular ad hoc networks (VANETs) (Härri et al., 2007) are fluctuating networks composed of a set of communicating vehicles (nodes) equipped with devices which are able to spontaneously interconnect each other without any pre-existing infrastructure. This means that no service provider is present in such kind of networks as it is usual in traditional or in mobile cellular communication networks. The most popular wireless networking technology available nowadays for establishing VANETs is the IEEE 802.11b WLAN, also known as WiFi (*wireless fidelity*). New standards such as the IEEE 802.11p and *WiFi direct* are promising but still not available to perform real tests with them. This implies that vehicles communicate within a limited range while moving, thus exhibiting a topology that may change quickly and in unpredictable ways. In such kind of networks, previous to its deployment, it is crucial to provide the user with an optimal configuration of the communication protocols in order to increase the effective data packet exchange, as well as to reduce the transmission time and the network use (with their implications on higher bandwidth and lower energy consumption). This is specially true in certain VANET scenarios (as shown in Fig. 1) in which buildings and distances discontinue communication channels frequently, and where the available time for connecting to vehicles could be just 1 s.

The efficient protocol configuration for VANETs without using automatic intelligent design tools is practically impossible because of the enormous number of possibilities. It is especially difficult (e.g., for a network designer) when considering multiple design issues, such as highly dynamic topologies and reduced coverage. In addition, the use of exact techniques is also impracticable due to the time spent during the great number of simulations required. All this motivates the use of metaheuristic techniques (Blum and Roli, 2003) which arise as well-suited tools to solve this kind of problems.

In this paper, we face the optimal File Transfer protocol Configuration (FTC) in VANETs by means of five different state-of-the-art optimization techniques. This problem lies in the core of any VANET application, and thus optimal configuration is a major concern. Also, we use many optimization algorithms because this is a new field, and their relative advantages are still unclear. Indeed, we cannot find results for comparisons in the literature since only manual (human expert) VDTP configurations were made so far. These algorithms are two swarm intelligence techniques: Particle Swarm Optimization (PSO) (Kennedy and Eberhart, 1995) and Differential Evolution (DE) (Price et al., 2005); two evolutionary algorithms: Genetic Algorithm (GA) (Blum and Roli, 2003) and Evolutionary Strategy (ES) (Beyer and Schwefel, 2002); and a trajectory search technique, Simulated Annealing (SA) Kirkpatrick et al. (1983). We have chosen these algorithms because they constitute a representative subset of well-known metaheuristics (population and trajectory based algorithms), with suitable operators for real parameter optimization, and with heterogeneous schemes of population and evolution. This way, we offer a set of initial results allowing future comparisons with other modern techniques.

* Corresponding author.
E-mail addresses: jnieto@lcc.uma.es (J. García-Nieto), jamal@lcc.uma.es (J. Toutouh), eat@lcc.uma.es (E. Alba).

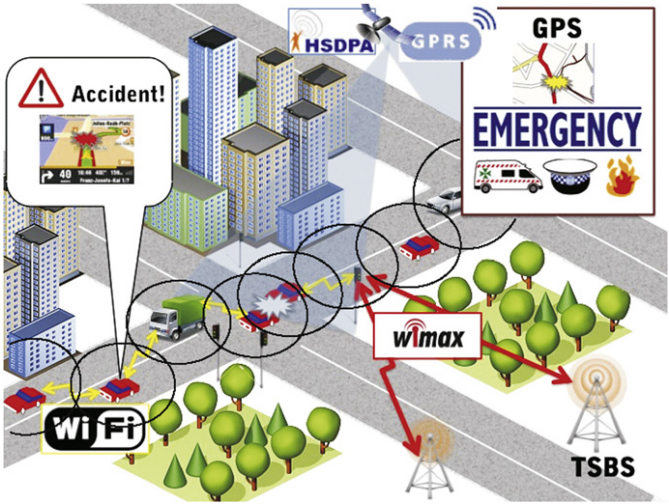

**Fig. 1.** Typical urban VANET scenario. Circles represent the WiFi coverage of vehicles.

For our tests, two typical car-to-car environment instances have been defined: Urban and Highway VANETs, both in special connection to the work done in the CARLINK CELTIC European Project for linking cars. We rely both on a flexible simulation structure using ns- 2 (The Network Simulator Project—Ns-2; Alba et al., 2008c) (a well-known realistic VANET simulator), and real tests for optimizing the transmission time, the number of lost packets, and the amount of data transferred. One additional contribution of this work is to provide the specialist with a useful platform, embedded within ns-2, to configure network protocols (available in http://neo.lcc.uma.es/staff/jamal/portal/) and hence obtaining a fair QoS control in VANETs.

The remaining of this paper is organized as follows. In the next section we briefly describe the most relevant related works found in the current literature. In Section 3 we introduce the Optimal File Transfer Configuration problem. Section 4 provides preliminary descriptions of the compared algorithms. In Section 5, the optimization strategy and fitness function are described. Experimental results and comparisons are presented in Section 6, including performance, scalability, and technical analyses of the resulted VANET configurations. Conclusions and future work are drawn in Section 7.

## 2. Related work

Few related works can be found in the specialized literature concerning the use of metaheuristics for the optimization of mobile ad hoc networks (MANETs). Vanhatupa et al. (2006) proposed a flexible Genetic Algorithm for optimizing channel assignment in mesh wireless networks. In that work, the network capacity was increased by 20% while keeping the coverage above 80%. In Alba et al. (2007c), a specialized Cellular Multi-Objective Genetic Algorithm (cMOGA) was used for finding an optimal broadcasting strategy in Urban MANETs, obtaining in this case three objectives fronts with coverage, bandwidth, and duration as performance metrics. The use of multi-objective techniques in this kind of works provides the specialists with a range of non-dominated solutions which can help them in the decision making process. Nevertheless, the use of (mono-objective) aggregated functions allows us the possibility of weighting the objectives and assign more (or less) importance to them for better guiding the search. This way, in Dorronsoro et al. (2008), six versions of GAs (panmictic and descentralized) were evaluated and successfully used in the design of ad hoc injection networks. From a different point of view, and due to its specific design, ant colony optimization (ACO) has been successfully adapted for implementing new routing protocols for MANETs (Di Caro et al., 2005), as well as for resource management (Chiang et al., 2007). Nevertheless, in these two last cases, the routing load provoked by the internal operations of ACOs makes these approaches unfeasible for large networks. More recently, Huang et al. (2009) proposed a new routing protocol based on a PSO to make scheduling decisions for reducing the packet loss rate in a theoretical VANET scenario.

In our work, besides of using the optimization technique itself as a protocol algorithm, our main contribution consists of improving the performance of an existing protocol by optimally tuning its parameters. This way, we will hopefully obtain optimal configurations in the network design phase without incorporating extra management load to the actual network operation.

## 3. Problem overview

The optimal File Transfer Configuration consists in optimizing the main parameters required by an application communication protocol. This protocol, called VDTP (vehicular data transfer protocol) (Alba et al., 2006), operates on the transport layer protocols of VANETs, allowing the *end-to-end* file transfer. This implies that considerations about the *multi-hop* interconnection mode and routing issues can be avoided, since they are carried out by the previous down layer protocols (e.g., UDP, DSR, IP, etc.). Therefore, the different vehicles that constitute the nodes in a given VANET can exchange complete files of information to each other by using VDTP. In this section, we briefly describe the VDTP, detailing the main parameters to be optimized.

### 3.1. Vehicular data transfer protocol

VDTP is a connectionless protocol which operates on DSR (Johnson et al., 2001), a routing protocol for multi-hop wireless ad hoc networks. In VDTP, the communication process is carried out by both a file *petitioner*, which tries to download a file, and a file *owner*, which stores the file. This transfer protocol operates by using the following packets: FIRQ (*file information request*), FIRP (*file information reply*), DRQ (*data request*), and DRP (*data reply*). As shown in Fig. 2(a), once the file petitioner knows the name and the location of a given file, it starts the communication by using the FIRQ packet in order to obtain the file size. Then, the petitioner waits for this information which is sent by the owner by means of a FIRP packet. After receiving the information about the file size, the petitioner computes the number of segments in which the file will be split, dividing the file size by the *chunk_size*. The petitioner starts the transfer by sending a DRQ(1) packet asking for the first segment of the file; then it waits for the first data chunk sent by the owner which uses the DRP(1) packet. This operation is repeated by both, petitioner and owner, until transferring the last chunk DRP($n$), and hence making up the complete file.

In VANETs, it is usual to work in a hostile medium which can provoke a high number of lost packets during the communication process. In this sense, VDTP provides the specialist with several mechanisms based on timers and counters, in order to solve such issues. The *timeout* mechanism controls the waiting time until a concrete DRQ or FIRQ packet has to be resent (*retransmission_time*). Fig. 2(b) shows an example of how the DRQ and the DRP packets are lost (and retransmitted) after an established timeout. The *counter* mechanism controls the number of DRQ/FIRQ packets that have been resent. As shown in Fig. 2(c), after a previously specified number of retransmissions (*total_attempts*) of the same DRQ/FIRQ packets, the communication between the vehicles is refused.

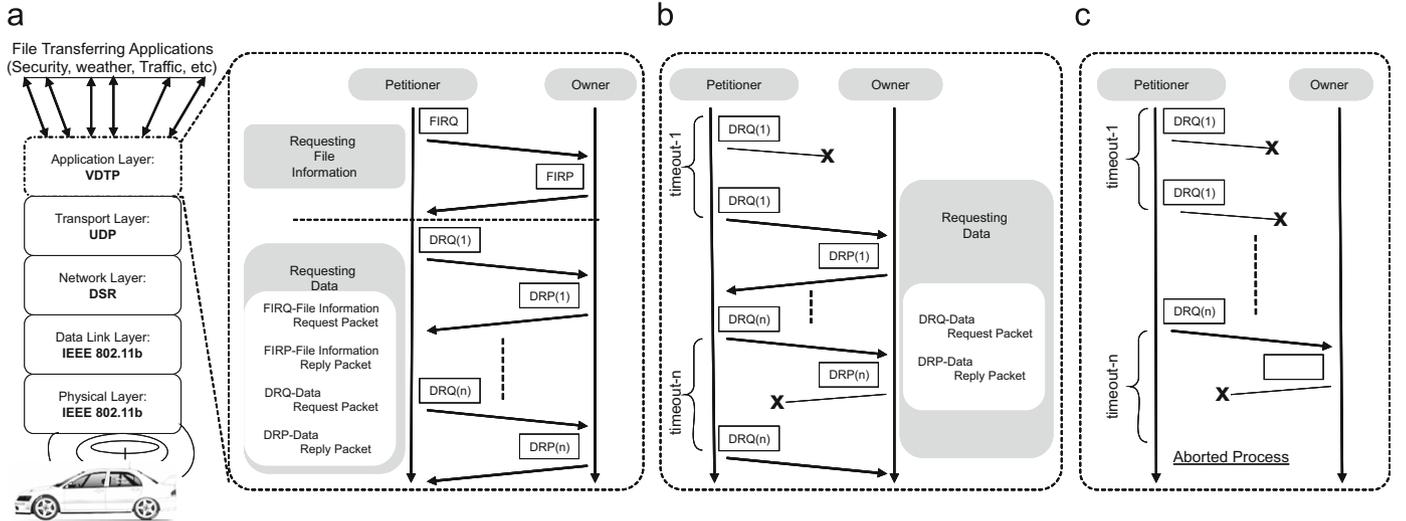

**Fig. 2.** VDTP operation modes: (a) a complete file exchange is done; (b) timeout expiration and retransmission; (c) communication refused.

### 3.1.1. Problem design variables

Since we are interested in finding the best possible configuration of VDTP, we have focused on the three aforementioned parameters: *chunk_size*, *retransmission_time* and number of *total_attempts*. Therefore, a given configuration (representing a solution of the problem) is a vector of three real values (*chunk_size*, *total_attempts* and *retransmission_time*). The range of each parameter is:

- *chunk_size*: $\mathbb{R}^+ \in [128 \cdots 524,288]$ bytes (524,288 bytes = 512 kBytes),
- *total_attempts*: $\mathbb{R}^+ \in [1 \cdots 250]$ attempts,
- *retransmission_time*: $\mathbb{R}^+ \in [1 \cdots 10]$ s.

These ranges were stated following the CARLINK consortium requirements for VANETs applications (http://carlink.lcc.uma.es).

## 4. The algorithms

In this section we briefly describe the five metaheuristic algorithms evaluated in this study. Specifically, they are two swarm intelligence techniques, Particle Swarm Optimization and Differential Evolution; two evolutionary algorithms, Genetic Algorithm and Evolutionary Strategy; and a trajectory search technique, Simulated Annealing. These techniques were selected with the aim of experimenting with different population structures, as well as different reproduction mechanisms. We have stated the same stop condition (reaching a certain number of generations) in all algorithms in order to simplify the following descriptions.

### 4.1. Particle Swarm Optimization (PSO)

Particle Swarm Optimization (Kennedy and Eberhart, 1995) is a population based metaheuristic inspired in the social behavior of birds within a flock, and initially designed for continuous optimization problems. In PSO, each potential solution to the problem is called *particle* and the population of particles is called *swarm*. In this algorithm, each particle position $x^i$ is updated each generation $g$ by means of the following equation:

$$x^i_{g+1} \leftarrow x^i_g + v^i_{g+1} \tag{1}$$

where factor $v^i_{g+1}$ is the velocity of the particle and is given by

$$v^i_{g+1} \leftarrow w \cdot v^i_g + \varphi_1 \cdot (p^i_g - x^i_g) + \varphi_2 \cdot (b_g - x^i_g) \tag{2}$$

In this formula, $p^i_g$ is the best solution that the particle $i$ has stored so far, $b_g$ is the best particle (also known as the *leader*) that the entire swarm has ever created, and $w$ is the inertia weight of the particle (it controls the trade-off between global and local experience). Finally, $\varphi_1$ and $\varphi_2$ are specific parameters which control the relative effect of the personal and global best particles ($\varphi_1 = \varphi_2 = 2 \cdot UN(0,1)$).

Algorithm 1 describes the pseudocode of PSO. The algorithm starts by initializing the swarm (Line 1), which includes both the positions and velocities of the particles. The corresponding $p^i$ of each particle is randomly initialized, as well as the leader $g$ (Line 2). Then, during a maximum number of iterations, each particle *flies* through the search space updating its velocity and position (Lines 5 and 6), it is then evaluated (Line 7), and its $p^i$ is also calculated (Lines 8). At the end of each iteration, the leader $b$ is updated.

**Algorithm 1.** Pseudocode of PSO.

```
1:    initializeSwarm()
2:    locateLeader(b)
3:    while g < maxGenerations do
4:      for each particle x^i_g do
5:        updateVelocity(v^i_g) //Equation 2
6:        updatePosition(x^i_g)// Equations 1
7:        evaluate(x^i_g)
8:        update(p^i_g)
9:      end for
10:     updateLeader(b_g)
11:   end while
```

## 4.2. Differential Evolution (DE)

Differential Evolution (Price et al., 2005) is a stochastic population based algorithm designed to solve optimization problems in continuous domains. The population consists of a set of individuals which evolve simultaneously through the search space of the problem. The task of generating new individuals is performed by differential operators such as the differential mutation and crossover. A *mutant individual* $w_{g+1}^i$ is generated by the following equation:

$$w_{g+1}^i \leftarrow v_g^{r1} + \mu \cdot (v_g^{r2} - v_g^{r3}) \quad (3)$$

where $r1, r2, r3 \in \{1, 2, \ldots, i-1, i+1, \ldots, N\}$ are random integers mutually different, and also different from the index $i$, the mutation constant $\mu > 0$ stands for the amplification of the difference between the individuals $v_g^{r2}$ and $v_g^{r3}$, and it avoids the stagnation of the search process.

In order to increase even more the diversity in the population, each mutated individual undergoes a crossover operation with the *target individual* $v_g^i$, by means of which a *trial individual* $u_{g+1}^i$ is generated. A randomly chosen position is taken from the mutant individual to prevent that the trial individual replicates the target individual.

$$u_{g+1}^i(j) \leftarrow \begin{cases} w_{g+1}^i(j) & \text{if } r(j) \leq Cr \text{ or } j = j_r \\ v_g^i(j) & \text{otherwise} \end{cases} \quad (4)$$

As shown in Eq. (4), the crossover operator randomly chooses a uniformly distributed integer value $j_r$ and a random real number $r \in (0, 1)$, also uniformly distributed for each component $j$ of the trial individual $u_{g+1}^i$. Then, the crossover probability $Cr$ and $r$ are compared just like $j$ and $j_r$. If $r$ is less than or equal to $Cr$ (or $j$ is equal to $j_r$) then we select the $j$ th element of the mutant individual to be allocated in the $j$ th element of the trial individual $u_{g+1}^i$. Otherwise, the $j$ th element of the target individual $v_g^i$ becomes the $j$ th element of the trial individual. Finally, a selection operator decides the acceptance of the trial individual for the next generation if and only if it yields a reduction in the value of the evaluation function (also called *fitness* function $f()$), as shown by the following equation:

$$v_{g+1}^i \leftarrow \begin{cases} u_{g+1}^i & \text{if } f(u_{g+1}^i) \leq f(v_g^i) \\ v_g^i(j) & \text{otherwise} \end{cases} \quad (5)$$

Algorithm 2 shows the pseudocode of DE. After initializing the population (Line 1), the individuals evolve during a number of generations (maxGenerations). Each individual is then mutated (Line 5) and recombined (Line 6). The new individual is selected (or not) following the operation of Eq. (5) (Lines 7 and 8).

**Algorithm 2.** Pseudocode of DE.

```
1:     initializePopulation()
2:     while g < maxGenerations do
3:       for each individual v_g^i do
4:         choose mutually different(r_1, r_2, r_3)
5:         w_{g+1}^i ← mutation(v_g^{r1}, v_g^{r2}, v_g^{r3}, μ)
6:         u_{g+1}^i ← crossover(v_g^i, w_{g+1}^i, cp)
7:         evaluate(u_{g+1}^i)
8:         v_{g+1}^i ← selection(v_g^i, u_{g+1}^i)
9:       end for
10:    end while
```

## 4.3. Genetic Algorithm (GA)

Genetic Algorithms (Blum and Roli, 2003) are the most popular metaheuristic algorithms. A GA iterates a process in which two parents are selected from the whole population with a given selection criterion, they are then recombined, the obtained offsprings are mutated, and finally they are evaluated and inserted back into the population following a given criterion. The mutation process is carried out by randomly (uniformly) selecting one of the elements in the solution, and assigning (randomly) a new value in the range as stated in Section 3.1.1. As recombination operator we use here a polynomial crossover defined for continuous variables (Blum and Roli, 2003). Algorithm 3 summarizes the operations of a canonical GA.

**Algorithm 3.** Pseudocode of GA.

```
1:     P_0 ← initializePopulation()
2:     while g < maxGenerations do
3:       P'_g ← recombine(P_g)
4:       P''_g ← mutate(P'_g)
5:       evaluate(P''_g)
6:       P_{g+1} ← select(P''_g ∪ P'_g)
7:     end while
```

There are two main versions of GA: *steady state* GA (ssGA) and *generational* GA (genGA). The difference between the ssGA and the genGA is the way in which the population is being updated with the new individuals generated during the evolution. In the case of the ssGA, new individuals are directly inserted into the current population while in the case of the genGA, a new auxiliary population is built with the obtained offsprings and then, once this auxiliary population is full, it completely replaces the current population. Thus, in ssGAs the population is asynchronously being updated with the newly generated individuals, while in the case of genGAs all the new individuals are updated at the same time, in a synchronous way.

## 4.4. Evolutionary Strategy (ES)

Evolutionary Strategy (Beyer and Schwefel, 2002) is a metaheuristic algorithm, designed by Rechenberg and Schwefel, also based on the ideas of adaptation and evolution.

As common with evolutionary algorithms, the mutation and selection operators are applied to the individuals through a given number of generations. The selection in evolutionary strategies is deterministic and only based on the fitness rankings, not on actual fitness values. We used here a mutation operator as explained in GA.

**Algorithm 4.** Pseudocode of ES.

```
1:     c_0 ← initializeParent()
2:     while g < maxGenerations do
3:       o_g ← mutate(c_g)
4:       evaluate(o_g)
5:       if f(o_g) is better than f(c_g) then
6:         c_g ← o_g
7:       end if
8:     end while
```

The canonical ES (Algorithm 4) operates on a population of size two: the current individual (parent $c$) and the result of its mutation (offspring $o$). After the parent initialization (Line 1), ES starts the evolution process by generating a mutated offspring (Line 3) which is evaluated (Line 4). Only if the offspring has a better fitness than the parent, it becomes the parent of the next generation (Lines 5 and 6). Otherwise the offspring is ignored. This is version of ES is called (1+1)-ES. More generally, in (1+λ)-ES, a population with more than one offsprings (λ) can be generated for being compared with the same parent. In a (1, $\lambda$)-ES the best offspring becomes the parent of the next generation while the current parent is always ignored. The most generalized version, ($\mu+/,\lambda$)- ES, often uses a population of parents ($\mu$) and also recombination as an additional operator.

### 4.5. Simulated Annealing (SA)

SA was first presented as a trajectory based optimization technique in Kirkpatrick et al. (1983). It is inspired in the metallurgy processes of annealing, and basically lies in a local search method with a mechanism that eventually promote solutions of worse quality than the current ones (uphill moves), in order to escape from local minima. The probability of performing such a movement decrease during the search process. The pseudocode of the canonical SA is showed in Algorithm 5.

**Algorithm 5.** Pseudocode of SA.

```
1:    initialize(T,S_a)
2:    evaluate(S_a)
3:    while g < maxGenerations do
4:      while not coolingCondition(g) do
5:        S_n ← chooseNeighbor(S_a)
6:        evaluate(S_n)
7:        if accept(S_a,S_N,T) then
8:          S_a←S_n
9:        end if
10:     end while
11:     coolDown(T)
12:   end while
```

The algorithm works iteratively keeping a single tentative solution $S_a$ at any time. In every iteration, a new solution $S_n$ is generated from the previous one, $S_a$ (Line 5), and either replaces it or not depending on an acceptance criterion (Lines 7–8). The acceptance criterion works as follows: both the old ($S_a$) and the new ($S_n$) solutions have an associated quality value, determined by a fitness function ($f()$). If it is worse, it replaces it with probability $prob$ (Eq. (6)). This probability depends on the difference between their quality values and control parameter $T$ named *temperature*. This acceptance criterion provides the way of escaping from local optima.

$$prob = \frac{2}{1+e^{(f(S_a)-f(S_n))/T}} \quad (6)$$

As iterations go on, the value of the temperature ($T$) is reduced following a cooling schedule (Line 11), thus biasing SA towards accepting only better solutions. In this work, we employ the geometric rule $T(n+1) = \alpha \cdot T(n)$, where $0 < \alpha < 1$, and the cooling is performed every $k$ iterations ($k$ is the Markov chain length).

For the neighbor selection, we use a mutation operator (as in GA and ES). The initial value of temperature $T$ is automatically generated in such a way that any movement from the initial (random) solution will be accepted with a certain probability.

## 5. Optimization strategy

Our optimization strategy for this problem is composed of basically two main parts: an optimization algorithm and a simulation procedure. The optimization part is carried out by (independently) one of the algorithms described in Section 4. All of them are specially adapted to find optimal (or cuasi-optimal) solutions in continuous search spaces (which is the case in this work). The simulation process is a way of assigning a quantitative quality value to the factors regulating VDTP, thus leading to optimal configurations of this protocol tailored to a given scenario. This procedure is carried out by means of the *ns-2* simulator in which we have implemented the VDTP protocol for sending files in VANETs.

In each optimization algorithm, the evaluation of each solution is carried out by means of the simulation component. As Fig. 3 illustrates, when a given algorithm generates a new solution it is immediately used for configuring the VDTP. This configuration evaluates the quality of the solution by using the received *retransmission time*, *chunk size*, and *total number of attempts*, as explained in Section 3.1. Then, *ns-2* is started and maps a given VANET scenario instance, taking its time in evaluating the scenario with buildings, signal loss, obstacles, vehicles, speed, covered area, etc., under the circumstances defined by the three control parameters optimized by the algorithm. After the simulation, *ns-2* returns the global information about the *transmission time* required for sending the file, the *number of lost packets* generated during the simulation, and the *amount of data* exchanged between vehicles. This information is used to compute the *fitness* function.

### 5.1. Fitness function

Since *ns-2* operates by simulating (and averaging) many potential variations scenario all fitting the actual vehicle system, there is a possibility of obtaining different fitness values even using the same VDTP configuration (solution). Therefore, in order to provide each solution with a fitness value as reliable as possible, a single evaluation of one solution requires $N=10$ internal simulations, computing the global fitness ($F$) as the mean

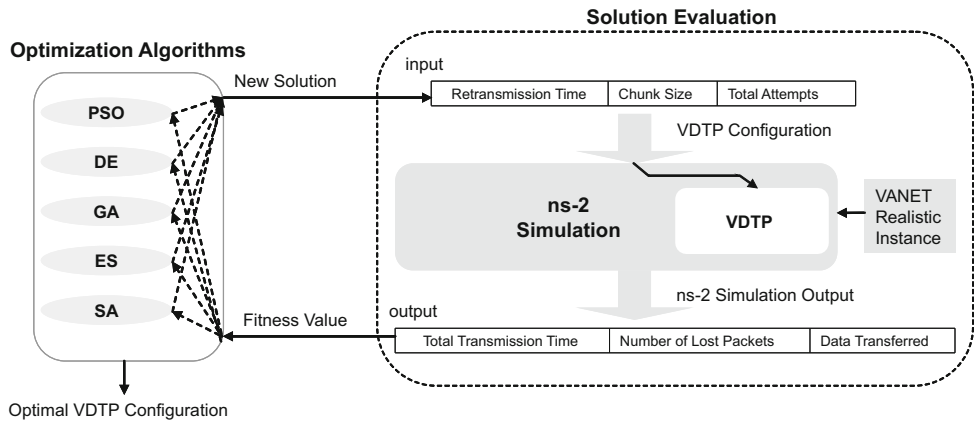

**Fig. 3.** Optimization strategy for VDTP configuration in VANETs. The algorithms invoke the ns-2 simulator each solution evaluation.

of all ns-2 results:

$$F = \frac{1}{N}\sum_{i=1}^{N} \frac{transmission\_time_i + lost\_packets_i}{\log(data\_transferred_i + C)} \quad (7)$$

In this equation, $i \in [1 \cdots 10]$ is the number of simulations per solution evaluation. The factor $C=2$ avoids division zero if there is no data transference, preventing a possible error in the fitness calculation. The data transferred is presented in logarithmic scale in order to make up for the difference in the range of values. This way, the algorithm looks for minimizing the global fitness.[1]

## 6. Experiments

We have used the implementation of the five algorithms provided by MALLBA (Alba et al., 2007a), a C++ based framework of metaheuristics for solving optimization problems. The simulation phase is carried out by running ns-2 simulator v-2.31. For the experiments, we made 30 independent runs of each algorithm on machines with Pentium IV 2.4 GHz core, 1 GB of RAM and O.S Linux Fedora core 6.

### 6.1. Instances: VANET scenarios

We have created two simulation VANET scenarios (instances) from real Urban and Highway areas of Málaga, Spain (selected areas in Fig. 4). These instances have been generated following the real tests carried out by experts in the scope of the CARLINK project, with the aim of obtaining as different as possible conditions of speed, number of vehicles, obstacles, signal noise, network use, etc. Therefore, we can analyze in both scenarios the behavior and performance of the compared algorithms, as well as the differences in the resulting VDTP configurations in terms of communication efficiency. Furthermore, we can compare these automatically generated configurations against the ones used in the real experiments by human experts in CARLINK (Alba et al., 2008a, 2008b).

#### 6.1.1. Urban

The Urban instance covers an area of 120,000 m² including buildings and semaphores. We have used VanetMobiSim (Alba et al., 2007b) for generating a realistic simulation mobility model where vehicles move randomly according to real traffic rules. A number of 30 vehicles move with a velocity between 30 and 50 km/h, and 20 of them trying to send and receive a file of 1024 kBytes.

#### 6.1.2. Highway

The Highway instance covers a stretch of 1 km with two directions without buildings and semaphores. In this case, the absence of obstacles is made up for the handicap of the high speed of vehicles, which also interferes the communication among vehicles. We have also used VanetMobiSim (Alba et al., 2007b) for generating a realistic simulation mobility model where vehicles move randomly according to real traffic rules. In the Highway VANET, a number of 30 vehicles move with a velocity between 80 and 110 km/h, and 20 of them trying to send and receive a file of 1024 kBytes size.

The resulted communication environments of Urban and Highway instances, including directions and mobile nodes (vehicles), were mapped in the ns-2 simulator following the VANET specifications of devices and protocols[2] summarized in Table 1. The ns-2 mobility trace definitions for both instances are publicly available in the following URL http://neo.lcc.uma.es/staff/jamal/portal/?q=content/malaga-scenario.

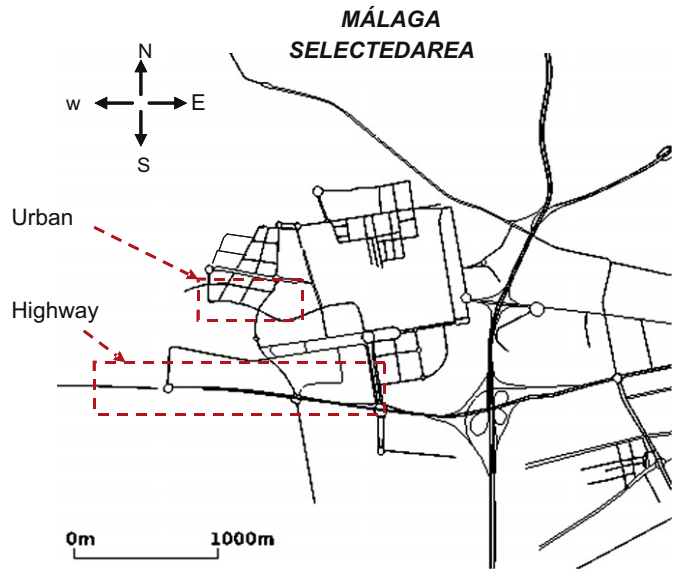

**Fig. 4.** Selected area map of Málaga for our VANET instances. Urban and Highway surfaces are enclosed by dotted lines.

**Table 1**
VANET instance specification.

| Parameter | Value |
| --- | --- |
| Propagation model | Two ray ground |
| Carrier frequency | 2.472 GHz |
| Channel bandwidth | 5.5 Mbps |
| Wifi channel | 13 |
| Link layer: transceiver | PROXIM ORiNOCO PCMCIA (IEEE 802.11b) |
| Link layer: antenna gain | 7 dBi (Omnidirectional) |
| Mac protocol | 802.11-b |
| Routing protocol | DSR |
| Transport protocol | UDP |
| Application protocol | VDTP |
| File transfers | 20 sessions |

### 6.2. Parameter settings

In our experiments, all studied algorithms were configured in order to perform 1000 solution evaluations per run. At each one of these solution evaluations, ns-2 performs 10 independent simulations of the target scenario with the same protocol configuration as stated in Section 5.1. Therefore, the population based algorithms (PSO, DE, GA, and $(\mu, \lambda)$-ES) were configured with 20 individuals, performing 50 generational steps.

Table 2 summarizes the remaining parameters specific to each algorithm. These parameters were selected as the most accurate after a set of initial tuning experiments. In these, a number of five combinations of parameters per algorithm and VANET instance were tested performing 10 independent runs per combination, hence resulting a number of 500 additional executions. Preliminary results of parameters tuning are available in Table A1 of Appendix A.

---

[1] A multi-objective evaluation (Deb, 2001) was not taken into account since objectives are not necessarily opposed in this work.

[2] DSR (dynamic source routing, Johnson et al., 2001), UDP (User Datagram Protocol), and VDTP (vehicular file transfer protocol).

**Table 2**
Parameterization of the optimization algorithms.

| Algorithm | Parameter | Symbol | Value |
|---|---|---|---|
| PSO | Local coefficient | $\varphi_1$ | $2 \cdot rand(0.1)$ |
| | Social coefficient | $\varphi_2$ | $2 \cdot rand(0.1)$ |
| | Inertia weigh | $w$ | 0.5 |
| DE | Crossover probability | $Cr$ | 0.9 |
| | Mutation factor | $\mu$ | 0.1 |
| GA | Crossover probability | $P_{cros}$ | 0.8 |
| | Mutation probability | $P_{mut}$ | 0.2 |
| ES | Crossover probability | $P_{cros}$ | 0.9 |
| | Mutation probability | $P_{mut}$ | 0.1 |
| SA | Temperature decay | $T$ | 0.8 |

**Table 3**
Final fitness values regarding the Urban and Highway VANET scenarios.

| Instance | Algorithm | Mean ± Std.dev. | Minimum | Median | Maximum |
|---|---|---|---|---|---|
| Urban | PSO | **1.6346 ± 0.2899** | 0.9077 | **1.7809** | **1.8918** |
| | DE | 1.7423 ± 0.3717 | **0.7389** | 1.8658 | 2.0228 |
| | GA | 1.9086 ± 0.2260 | 0.8799 | 1.9731 | 2.1614 |
| | ES | 2.1517 ± 0.1266 | 1.8862 | 2.1222 | 2.4246 |
| | SA | 2.7850 ± 0.8718 | 0.8730 | 2.1663 | 3.8025 |
| Highway | PSO | **4.1761 ± 0.2556** | 3.3301 | 4.2513 | **4.4554** |
| | DE | 4.6631 ± 0.9328 | 2.7145 | 4.2272 | 7.0531 |
| | GA | 4.3805 ± 0.8695 | **2.5345** | 4.1918 | 5.8608 |
| | ES | 5.7833 ± 0.9705 | 3.8836 | 6.1347 | 6.9421 |
| | SA | 4.4246 ± 0.7401 | 3.1498 | **4.0855** | 5.7922 |

Columns 3 contains the mean and standard deviation (Std. dev.) of the fitness values in 30 independent runs. Columns 4, 5, and 6 show the minimum, median, and maximum values of fitness, respectively.

### 6.3. Results and comparisons

In this section we present the results obtained by the five studied algorithms when solving the optimal File Transfer Configuration (FTC) problem on VDTP. Table 3 shows the resulting fitness values regarding the Urban and Highway VANET scenarios in terms of the mean, the standard deviation, the minimum (best fitness), the median, and the maximum (worst fitness) found in 30 independent runs of every algorithm.

For the Urban scenario, we can observe (in Table 3) that PSO obtained the best result in terms of the mean fitness. This smallest mean value leads us to believe that using the PSO the resulting VDTP ends in an efficient communication which is fast and accurate between vehicles. In addition, the best median and maximum values were also obtained by PSO, although the best minimum (e.g., the best VDTP configuration found for Urban) was reached by DE. This is an expected value, since DE generally shows a pronounced exploitative behavior (using a parametrization close to the standard one, Price et al., 2005), while PSO tends to have an explorative performance using a high inertia (as in this study $w$=0.5, Eberhart and Shi, 2000). Similar results can be observed for the Highway scenario, in which PSO obtained the best mean fitness value again. For this instance, PSO also showed the lowest value of standard deviation. This implies a considerable advantage, since it provides our model with a high robustness, which is a crucial issue when designing VANETs. In terms of the minimum fitness, GA and DE obtained the best VDTP configurations for the Highway scenario. The worst configuration was obtained by ES.

**Table 4**
PSO versus other algorithms Signed Rank test with confidence level 95% ($p$-value=0.05).

| Algorithm | Urban | | Highway | |
|---|---|---|---|---|
| | Test | $p$-value | Test | $p$-value |
| DE | ▲ | 0.047 | ▲ | 0.001 |
| GA | ▲ | 0.001 | △ | 0.453 |
| ES | ▲ | 0.001 | ▲ | 0.001 |
| SA | ▲ | 0.001 | △ | 0.371 |

**Table 5**
Friedman rank test with confidence level 95%.

| Urban | | Highway | |
|---|---|---|---|
| Algorithm | Rank | Algorithm | Rank |
| PSO | **1.27** | SA | **1.83** |
| DE | 1.83 | GA | 1.97 |
| GA | 3.07 | PSO | 2.17 |
| ES | 4.33 | DE | 3.67 |
| SA | 4.50 | ES | 4.97 |

In order to provide such comparison with statistical meaning, we have applied a Signed Rank (Wilcox, 1987) statistical test to the distributions of the aforementioned results. We have used this non-parametric[3] test with confidence level of 95% ($p$-value=0.05), which leads us to ensure that these results are statistically different if they result in $p$-value < 0.05. Table 4 contains the resulted $p$-value of applying the Signed Rank test to PSO (the one with the best mean fitness) in comparison with the remaining of algorithms, hence confirming the differences in results. In this table, the symbol ▲ means that PSO is statistically better than the compared algorithm, whereas the symbol △ means that PSO has a better rank than the compared algorithm, but without statistical difference.

As we can observe in Table 4, PSO is statistically better than all compared algorithms for the Urban instance. Only DE shows a $p$-value (0.047) close to 0.05, being lower in any case. Concerning the Highway instance, PSO shows the best rank, not far from GA and SA.

A general comparison can be made using the Friedman (Iman and Davenport, 1980) statistical test by means of which the algorithms are sorted in a ranked list. Table 5 shows the Friedman ranking of the compared algorithms in Urban and Highway instances (the best ranked algorithm is in the top). For Urban instance, PSO and DE are the best ranked algorithms, but showing SA the last position. Nevertheless, for Highway scenario, SA obtains the best rank, whereas PSO is located in the third position.

Theses statistical results lead us to think that, in spite of the global best behavior of PSO, the different requirements implicit to both instances implies that each algorithm can show quite different results depending on the VANET scenario on which it operates. For example, DE shows a competitive performance in Urban scenario whereas it is the second worst in Highway. The opposite example can be observed in GA and SA which show weak results in Urban but highly competitive ones in Highway. Therefore, the VANET designer can select the optimization model

---
[3] The distributions violate the condition of normality required to apply parametric tests (Z Kolmogorov–Smirnov = 0.009).

more suited to his/her requirements, and choose the best option for each studied VANET scenario.

### 6.4. Performance analysis

We present now a performance study which basically lies in analyzing the best fitness value, resulted from each function evaluation, during the whole evolution process of a given algorithm. Figs. 5 and 6 illustrate the graphs of the best fitness values (communication cost) obtained through the median execution in Urban and Highway instances, respectively.

We can observe in both graphics that PSO and DE tend to converge in the same range of solution evaluations, although they could improved their fitness even in the final steps of the evolution process. GA shows a similar trend as the former ones but it is subjected to an early stagnation.

Finally, the different behaviors observed in ES, and specifically in SA, for Urban and Highway instances confirm us the high dependency of such algorithms to each different VANET instance (they are not robust in this application).

Concerning the mean run time that each algorithm spent in the experiments, Table 6 shows both the mean time in which the best solution was found $T_{best}$, and the global mean run time $T_{run}$ for Urban and Highway scenarios. In general, SA shows the shortest times to find the best solution for the two VANET instances. We suspect that despite its temperature mechanism, SA quickly falls in local optima hence obtaining weak results in Urban scenario. Nevertheless, this behavior can be an advantage for Highway scenario where SA obtained accurate solutions with a fast performance. As expected in PSO and DE, they spent closed executions times for the two VANET instances since they have similar internal operations. This resemblance in time consumption was also registered in the two evolutionary algorithms, GA and ES.

As a summary, the algorithms use between 9.00E+03 and 4.76E+03 s for the Urban scenario (150 and 80 minutes, respectively), and between 2.19E+03 and 8.45E+02 s for Highway scenario (60 and 23 minutes, respectively). This relative low effort in the protocol design is completely justified by the subsequent benefits obtained in the global data transmission time and loss of packets once the VANET is physically deployed as observed in the following analysis.

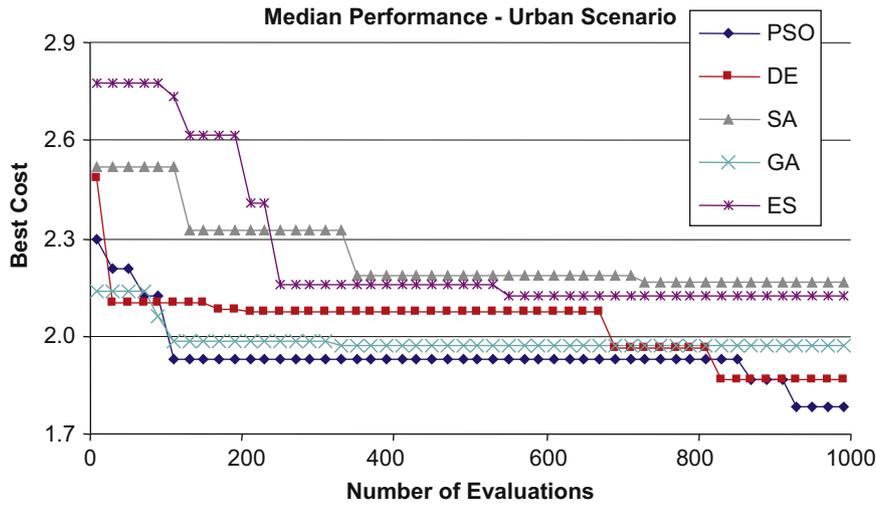

Fig. 5. Median performance in Urban scenario.

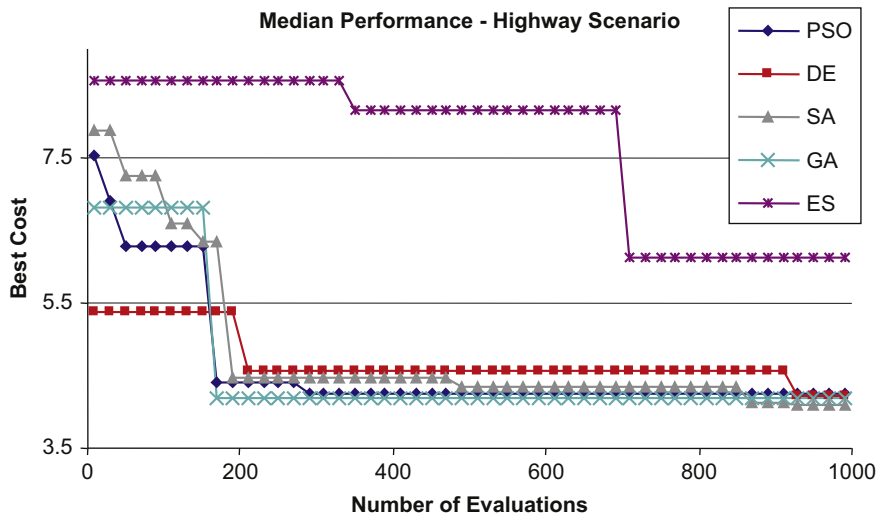

Fig. 6. Median performance in Highway scenario.

### 6.5. Scalability analysis

Once we have analyzed the performance of the five algorithms in two different VANET scenarios, we study in this section how do various network sizes affect the performance of these optimization techniques. For this purpose, we have generated two new VANET instances from the initial Urban scenario (of Málaga) by enlarging the metropolitan area considered. Therefore, as Fig. 7 shows, the initial urban area (A1) has been expanded to A2 and A3 VANET areas. We have set the traffic flow as described in Section 6.1, also increasing the number of vehicles as follows:

- $Urban_{A\ 1}$ with 30 vehicles in 120,000 m$^2$,
- $Urban_{A\ 2}$ with 40 vehicles in 240,000 m$^2$,
- $Urban_{A\ 3}$ with 50 vehicles in 360,000 m$^2$.

From the point of view of the mean fitness obtained by each algorithm (out of 30 independent runs), we can observe in Table 7 that PSO keeps the best performance for $Urban_{A\ 2}$ and $Urban_{A\ 3}$.

Additionally, one of the most interesting results can be observed in GA, which arises as the second best algorithm in improving its behavior with the VANET size. ES obtains moderate mean fitness values for all network instances, keeping a low standard deviation. The worst results are registered by SA in $Urban_{A\ 2}$, and DE in $Urban_{A\ 3}$. Concerning DE, the initial choice of its parameters ($Cr=0.9$ and $\mu=0.1$) could lead the algorithm to perform an exploitative search, hence obtaining good results in small instances (the second best for $Urban_{A\ 1}$) but damaging its behavior in larger VANETs (the worst for $Urban_{A\ 3}$). In summary, excepting for GA and DE, we can confirm that for the scaled VANET instances the performance of the algorithms are similar to their performances in $Urban_{A\ 1}$ (the initial Urban VANET instance) being PSO always the best procedure.

A secondary but also interesting observation lies in the mean fitness values, which are in $Urban_{A\ 2}$ lower than in $Urban_{A\ 1}$. We suspect that, in spite of the larger dimension of $Urban_{A\ 2}$, the proportion of communicating vehicles (per m$^2$) in this VANET helps the protocol operation specially for intermediate nodes, hence improving the effective ratio of delivery packets and the overall retransmission time. This proportion could not be enough for $Urban_{A\ 3}$ where the cost of transmissions is the larger one.

Concerning the execution time, Table 7 shows in the three last columns the time required to find the best solution ($T_{best}$) for each VANET instance. Surprisingly, for PSO, ES, and SA the time required to converge in $Urban_{A\ 2}$ is lower than in $Urban_{A\ 1}$. This behavior can be explained by the fact of obtaining good solutions faster in $Urban_{A\ 2}$ than in $Urban_{A\ 1}$, where the lower number of vehicles could harm the communications conditions. On the contrary, the global run time ($T_{run}$) always increases with the network size. This is of course an expected result.

### 6.6. QoS analysis

Finally, from the point of view of the worked VDTP configurations (solutions), we analyze the results in terms of the QoS indicators considered here: the transmission time, the number of lost packets, and the amount of data transferred induced in the designed VANET. In this sense, Table 8 shows the results after simulating the best solutions found by the studied algorithms. In addition, the last row of this table contains the results of simulating the configuration of VDTP that has been used in the scope of the CARLINK project (real word results with actual cars).

For the Urban VANET, the VDTP configuration obtained by PSO (Chunk_Size=41,358 Bytes, Retransmission_Time=10 s, and num-

**Table 6**
Mean execution time (seconds) per independent run of each algorithm for Urban and Highway scenarios.

| Instance | Algorithm | $T_{best}$ (seconds) | $T_{run}$ (seconds) |
|---|---|---|---|
| Urban | PSO | 4.68E+03 | 7.95E+03 |
| | DE | 4.37E+03 | 7.12E+03 |
| | GA | 3.48E+03 | 6.68E+03 |
| | ES | 5.46E+03 | 9.00E+03 |
| | SA | **2.18E+03** | **4.76E+03** |
| Highway | PSO | 1.39E+03 | 2.19E+03 |
| | DE | 9.82E+02 | 2.10E+03 |
| | GA | 8.83E+02 | 1.56E+03 |
| | ES | 9.84E+02 | 1.47E+03 |
| | SA | **5.85E+02** | **8.45E+02** |

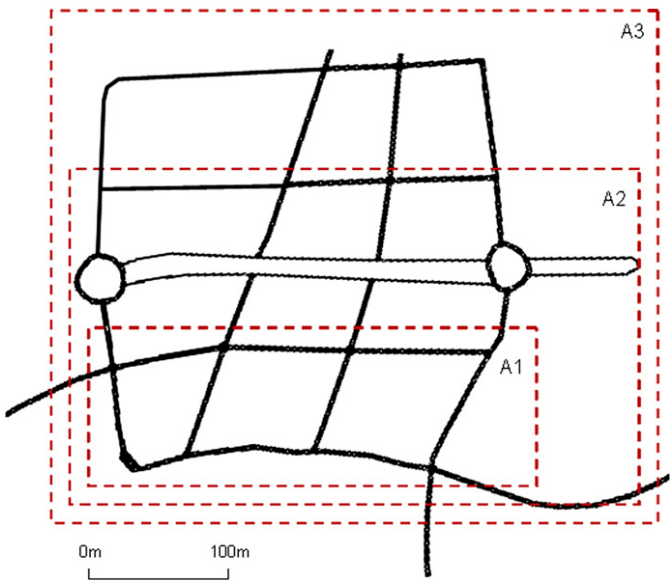

**Fig. 7.** Three urban areas from Málaga. Each area conforms a VANET instance.

**Table 7**
Performance comparison in terms of mean fitness and mean optimization time ($T_{best}$) of the three scaled Urban VANETs.

| Algorithm | Mean fitness | | | $T_{best}$ | | |
|---|---|---|---|---|---|---|
| | $Urban_{A\ 1}$ | $Urban_{A\ 2}$ | $Urban_{A\ 3}$ | $Urban_{A\ 1}$ | $Urban_{A\ 2}$ | $Urban_{A\ 3}$ |
| PSO | **1.6346 ± 0.2899** | **1.3920 ± 0.2831** | **3.6763 ± 0.4435** | 7.95E+03 | 5.93E+03 | 1.20E+04 |
| DE | 1.7423 ± 0.3717 | 1.4504 ± 0.1885 | 3.9186 ± 0.7419 | 7.12E+03 | 1.10E+04 | 1.43E+04 |
| GA | 1.9086 ± 0.2260 | 1.4100 ± 0.1235 | 3.6829 ± 0.5063 | 6.68E+03 | 9.81E+03 | 1.41E+04 |
| ES | 2.1517 ± 0.1266 | 1.5462 ± 0.6023 | 3.7799 ± 0.6227 | 9.00E+03 | 8.99E+03 | 1.50E+04 |
| SA | 2.7850 ± 0.8718 | 2.3880 ± 1.0207 | 3.8143 ± 0.1260 | **4.76E+03** | **3.40E+03** | **5.36E+03** |

**Table 8**
VDTP configurations and simulation output values for the optimal fitness achieved (in the median execution) by all studied algorithms.

| Instance | Algorithm | VDTP configuration | | | Simulation results | | |
|---|---|---|---|---|---|---|---|
| | | Chunk size (Bytes) | Retrans. time (s) | Attempts | Trans. Time (s) | Lost Packets | Data Transferred (kBytes) |
| Urban | PSO | 41.358 | 10.00 | 3 | **3.41** | **0.27** | 1.024 |
| | DE | 28.278 | 6.00 | 9 | 3.59 | 0.63 | 1.024 |
| | GA | 31.196 | 3.83 | 9 | 3.61 | **0.27** | 1.024 |
| | ES | 23.433 | 10.00 | 8 | 3.50 | **0.27** | 1.024 |
| | SA | 19.756 | 6.43 | 3 | 4.22 | 0.36 | 1.024 |
| | Human experts | 25.600 | 8.00 | 8 | 4.24 | 1.60 | 1.024 |
| Highway | PSO | 29.257 | 6.42 | 9 | **24.67** | 3.18 | 1.024 |
| | DE | 19.810 | 6.91 | 8 | 27.66 | 3.45 | 1.024 |
| | GA | 34.542 | 9.54 | 10 | 26.96 | 2.72 | 1.024 |
| | ES | 38.490 | 8.15 | 12 | 33.99 | 3.36 | 1.024 |
| | SA | 32.002 | 8.21 | 4 | 25.43 | **2.54** | 1.024 |
| | Human experts | 25.600 | 10.00 | 10 | 33.08 | 3.27 | 1.024 |

The last row contains the results obtained in the scope of the CARLINK project.

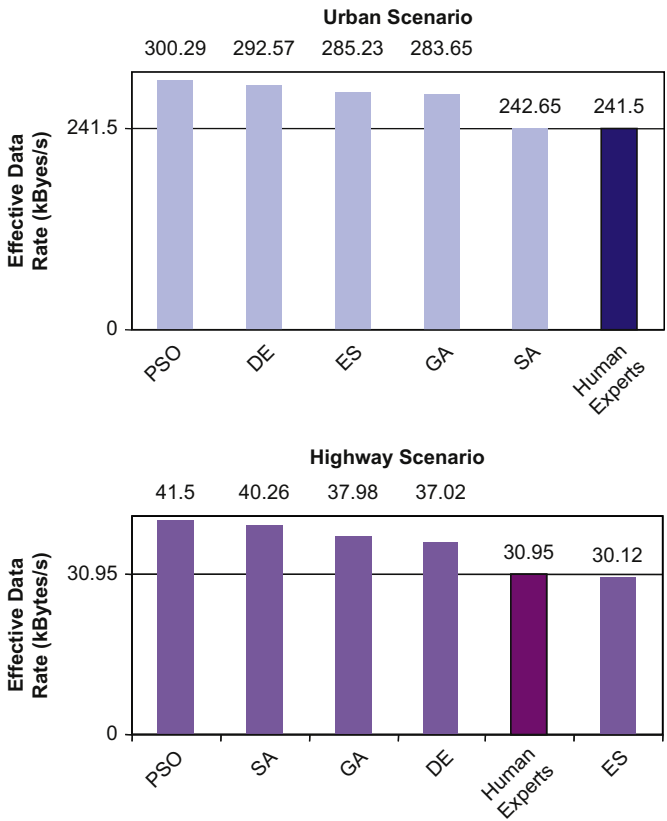

**Fig. 8.** Effective transmission data rates (throughput) (kBytes/s) achieved during the simulations of the final VDTP configurations in comparison with real car values given by human expert configurations (CARLINK).

ber of Attempts=3) achieves the best performance in terms of transmission time and mean number of lost packets. Specifically, in comparison with the human experts configuration of CARLINK, PSO obtains a reduction in the transmission time of 0.83 s (19.5%) registering also a lower number of lost packets.

Nevertheless, it is in the Highway scenario were PSO obtains the higher time reduction of 8.41 s (25%) regarding the human experts configuration (from 33.08 to 24.67 s). We must notice that, in spite of achieving the PSO a higher reduction in the transmission time than SA and GA, the fact of losing more packets (3.18 in PSO, 2.71 in GA and 2.54 in SA) in the global transference leads SA and GA to calculate a better fitness value (as shown in Table 3).

A final analysis can be done concerning one main QoS indicator: the *effective transmission data rate (throughput)*[4] achieved. As we can observe in Fig. 8, the VDTP configuration obtained by practically all algorithms in the two VANET scenarios obtained higher effective data rates than the human configured VDTP. Specifically, PSO achieves the highest effective data rate (300.29 kBytes/s in Urban and 41.54 kBytes/s in Highway). This clearly claims for the utilization of these automatic algorithms to help human designers. We again remind that the actual correction of effective data rates between cars are in the order of tens of kBytes/s, so our savings (58.79 kBytes/s in Urban and 10.5 kBytes/s in Highway) are truly meaningful in current real applications such as safety, traffic control, and weather predictions.

## 7. Conclusions

In this paper, we tackle the optimal File Transfer protocol Configuration (FTC) in VANETs by means of five popular metaheuristic algorithms. For this, we need a complex system accounting for a flexible simulation structure targeted for optimizing the transmission time, the number of lost packets, and the amount of data transferred in simulated and also realistic VANET scenarios.

The experiments, using *ns*-2 (well-known VANET simulator), reveal that all algorithms are capable of efficiently solve the optimum FTC problem. In the comparisons, PSO performs statistically better than all algorithms in Urban and statistically better than DE and ES in Highway. In addition, GA and SA show a competitive performance in Highway. The scalability analysis shows that GA improves with the network size, whereas DE decreases its performance with large VANET instances. PSO keeps the best result even for larger instances.

From the point of view of its real world utilization, PSO can reduce 19% of the transmission time in Urban and 25.43% in Highway with regards to human experts configuration of CARLINK, while transmitting the same amount of data (1024

---
[4] In our fitness function, instead of using the *throughput* as extra control parameter, we have broken down it into the transmission time and data transferred directly in order to count them separately and enhance the search process of the algorithms.

**Table A1**
Different combinations and results of the preliminary parameter tuning.

| Algorithm | Parameter Instances | Values Results | | | | |
|---|---|---|---|---|---|---|
| PSO | $\varphi_1$ | 2.0 | 2.0 | 2.0 | 2.0 | 2.0 |
| | $\varphi_2$ | 2.0 | 2.0 | 2.0 | 2.0 | 2.0 |
| | $w$ | 0.1 | 0.3 | 0.5 | 0.7 | 0.9 |
| | Urban | 1.952 | 1.978 | **1.634** | 2.766 | 3.280 |
| | Highway | 5.676 | 4.622 | **4.1761** | 5.283 | 6.045 |
| DE | $Cr$ | 0.1 | 0.3 | 0.5 | 0.7 | 0.9 |
| | $\mu$ | 0.9 | 0.7 | 0.5 | 0.3 | 0.1 |
| | Urban | 4.027 | 2.647 | 2.241 | 1.866 | **1.742** |
| | Highway | 7.255 | 5.622 | 4.776 | 4.734 | **4.663** |
| GA | $P_{cros}$ | 0.2 | 0.4 | 0.6 | 0.8 | 1.0 |
| | $P_{mut}$ | 0.8 | 0.6 | 0.4 | 0.2 | 0.1 |
| | Urban | 2.701 | 2.245 | 1.953 | **1.908** | 2.077 |
| | Highway | 5.216 | 4.848 | **4.380** | 4.490 | 4.609 |
| ES | $P_{cros}$ | 0.1 | 0.3 | 0.5 | 0.7 | 0.9 |
| | $P_{mut}$ | 0.9 | 0.7 | 0.5 | 0.3 | 0.1 |
| | Urban | 4.920 | 3.878 | 3.031 | 2.606 | **2.151** |
| | Highway | 7.836 | 6.877 | 6.240 | **5.783** | 5.923 |
| SA | $T$ | 0.2 | 0.4 | 0.6 | 0.8 | 1.0 |
| | Urban | 4.922 | 1.978 | 2.785 | **1.634** | 3.744 |
| | Highway | 7.665 | 5.201 | 4.820 | **4.424** | 4.683 |

kBytes). The highest effective data rates obtained by PSO (of 300.39 kBytes/s in comparison with 241.5 kBytes/s of human experts) and DE (292.57 kBytes/s) in Urban lead us to advise the final use of our automatic design algorithms.

As a matter of further work we are presently extending our benchmark with new VANET realistic instances (e.g., complete cities and highway knots). In addition, we are planning to define new optimized configuration schemes for other communication protocols such as: UDP, DSR, etc. which should efficiently support actual VANET design.

## Acknowledgments

Authors acknowledge funds from the Spanish Ministry MICINN and FEDER under contract TIN2008-06491-C04-01 (M* http://mstar.lcc.uma.es) and CICE, Junta de Andalucía, under contract P07-TIC-03044 (DIRICOM http://diricom.lcc.uma.es). José García-Nieto is supported by Grant BES-2009-018767 from the MICINN.

## Appendix A. Parameters tuning

Table A1 shows the results obtained in the preliminary parameters tuning procedure. A number of five combinations of parameters per algorithm and VANET instance were tested performing 10 independent runs per combination, hence resulting a number of 500 executions.